\documentclass[10pt,twocolumn,letterpaper]{article}

\usepackage{cvpr}
\usepackage{times}
\usepackage{epsfig}
\usepackage{graphicx}
\usepackage{amsmath}
\usepackage{amssymb}

\usepackage{multirow}
\usepackage{booktabs}
\usepackage[inline]{enumitem}

\usepackage[breaklinks=true,bookmarks=false]{hyperref}

\cvprfinalcopy 

\def\cvprPaperID{****} 

\begin{document}
\title{Light-Head R-CNN: In Defense of Two-Stage Object Detector}

\author{Zeming Li$^1$, Chao Peng$^2$, Gang Yu$^2$, Xiangyu Zhang$^2$, Yangdong Deng$^1$, Jian Sun$^2$  \\ $^1$School of Software, Tsinghua University,  \{lizm15@mails.tsinghua.edu.cn, dengyd@tsinghua.edu.cn \}  \\$^2$
Megvii Inc. (Face++), \{pengchao, yugang, zhangxiangyu, sunjian\}@megvii.com }

\maketitle

\begin{abstract}

In this paper, we first investigate why typical two-stage methods are not as fast as single-stage, fast detectors like YOLO~\cite{yolo,yolo9000} and SSD~\cite{ssd}. We find that Faster R-CNN~\cite{faster_rcnn} and R-FCN~\cite{rfcn} perform an intensive computation after or before RoI warping. Faster R-CNN involves two fully connected layers for RoI recognition, while R-FCN produces a large score maps. Thus, the speed of these networks is slow due to the heavy-head design in the architecture. Even if we significantly reduce the base model, the computation cost cannot be largely decreased accordingly. 

We propose a new two-stage detector, Light-Head R-CNN, to address the shortcoming in current two-stage approaches. In our design, we make the head of network as light as possible, by using a thin feature map and a cheap R-CNN subnet (pooling and single fully-connected layer).  Our ResNet-101 based light-head R-CNN outperforms state-of-art object detectors on COCO while keeping time efficiency. More importantly, simply replacing the backbone with a tiny network (e.g, Xception), our Light-Head R-CNN gets 30.7 mmAP at 102 FPS on COCO, significantly outperforming the single-stage, fast detectors like YOLO~\cite{yolo,yolo9000} and SSD~\cite{ssd} on both speed and accuracy. Code will be made publicly available.
\end{abstract}

\section{Introduction}  \label{sec:introduction}
\begin{figure}[h]
   \begin{center}
      \includegraphics[clip=true, ,width=1\linewidth]{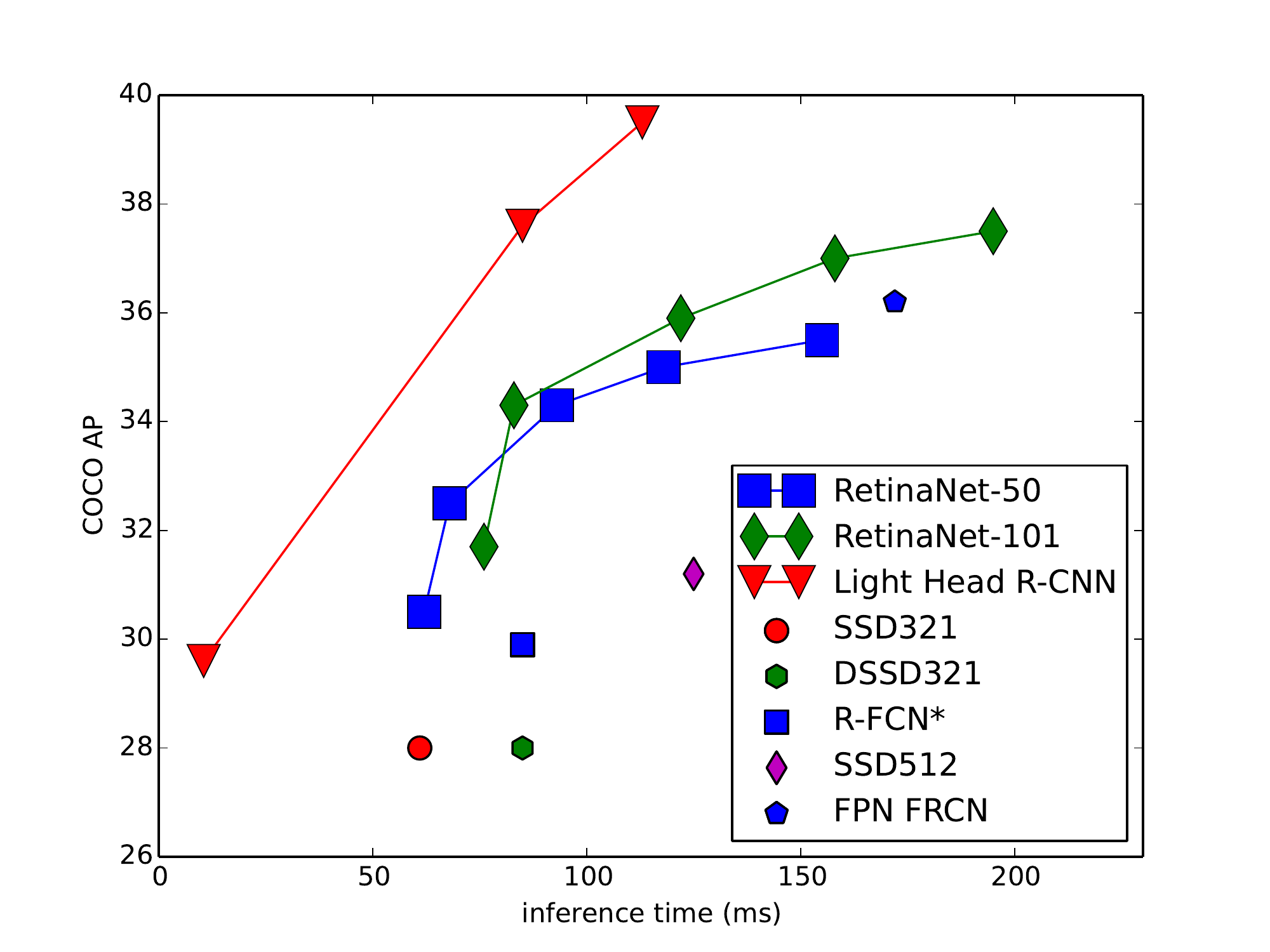}
   \end{center}
   \caption{Comparisons of Light Head R-CNN along with previous one-stage and two-stage detectors. We show our results with different backbones~(a small Xception like network, Resnet-50, Resnet-101). Thanks for better design principle, our Light Head R-CNN significant outperform all competitors, and provide a new upper envelope. Note that all results reported here are obtained by use single-scale training only. The multi-scale training results are presented in Table \ref{table:COCO_methods_comparison}.}
   \label{fig: intro_figure}
\end{figure}

Recent CNN-based object detectors can be categorized into single-stage detectors~\cite{yolo,yolo9000,ssd,focal_loss,dssd} and two-stage~\cite{fast_rcnn,faster_rcnn,fpn,mask_rcnn} detectors. The single-stage detector usually targets on a sweet-spot of very fast speed and reasonably good accuracy. The two-stage detector divides the task into two steps: the first step (\emph{body}) generates many proposals, and the second step (\emph{head}) focuses on the recognition of the proposals. Usually,  in order to achieve the best accuracy, the design of the head is heavy. The two-stage detector often has a sweet-spot of (relatively) slow speed and very high accuracy.   

Could the two-stage detector beat the single-stage detector on both efficiency and accuracy? We find that typical two-stage object detectors like Faster R-CNN~\cite{fast_rcnn} and R-FCN~\cite{rfcn} share the similar characteristics: a heavy head attached to the backbone network. For example, Faster R-CNN employs two large fully connected layers or all the convolution layers in ResNet 5-$th$ stage~\cite{faster_rcnn, NOC} for per RoI’s recognition and regression. It is time-consuming in terms of per-region prediction and even gets worse when a large number of proposals are utilized. In addition, the number of feature channels after ROI pooling is large, which makes the first fully connected consume a large memory and potentially influence the computational speed. Different from Fast/Faster R-CNN which apply a per-region subnetwork many times, Region-based fully convolutional networks (R-FCN)~\cite{rfcn} tries to share computation across all RoIs. However, R-FCN needs to produce a very large additional score maps with~$\#classes \times p \times p$~($p$ is the followed pooling size) channels, which is also memory and time consuming. The heavy head design of Faster R-CNN or R-FCN make the two-stage approach less competitive if we use a small backbone network.

In this paper, we propose a light-head design to build an efficient yet accurate two-stage detector. Specifically, we apply a large-kernel separable convolution to produce ``thin" feature maps with small channel number~($\alpha \times p \times p$ is used in our experiments and $\alpha \leq 10$). This design greatly reduces the computation of following RoI-wise subnetwork and makes the detection system memory-friendly. A cheap single fully-connected layer is attached to the pooling layer, which well exploits the feature representation for classification and regression. 

Because of our light-head structure, our detector is able to strike the best tradeoff of speed and accuracy, not matter a large or small backbone network is used.  As shown in Figure~\ref{fig: intro_figure}, our algorithm, dotted as Light-Head R-CNN, can significantly outperform the fast single-stage detector like SSD~\cite{ssd} and YOLOv2~\cite{yolo9000} with even faster computational speed. In addition, our algorithm is also flexible to large backbone network. Based on a ResNet-101 backbone, we can outperform state-of-art algorithms including two-stage detectors like Mask R-CNN~\cite{mask_rcnn} and one-stage detectors like RetinaNet~\cite{focal_loss}.

\section{Related works} 
Benefited from the rapid development of deep convolutional networks\cite{alexnet,vggnet,googlenet,he2016deep,bath_norm,resnext,senet,shufflenet,senet,squeezenet,mobilenet}, a great progress has been made for the object detection problem. We briefly review some of the recent object detection work in two dimensions as follows: 

\textbf{Accuracy perspective:}  
R-CNN~\cite{girshick2014rcnn} is among the first to utilize deep neural network features into detection system. Hand-engineered methods, such as Selective Search~\cite{selective_search}, Edge Boxes~\cite{edge_boxes}, MCG\cite{multiscale_com_group}, is involved to generate proposals for R-CNN. Then Fast R-CNN~\cite{fast_rcnn} is proposed to join train object classification and bounding box regression, which improves the performance by multi-task training. Following Fast R-CNN, Faster R-CNN~\cite{faster_rcnn} introduces Region Proposal Network~(RPN) to generate proposals by using network features. Benefited from richer proposals, it marginally increase the accuracy. Faster R-CNN has been seen as a milestone of R-CNN serials detectors. Most of the following works strengthen Faster R-CNN by bringing more computation into network. Dai \emph{et al.} propose Deformable Convolutional Networks~\cite{deformable} to model geometric transformations by learning additional offsets without supervision. Lin \emph{et al.} propose Feature Pyramid Networks~(FPN)~\cite{fpn}, which exploits inherent multi-scale, pyramidal hierarchy of deep convolutional networks to construct feature pyramids. Based on FPN, Mask R-CNN~\cite{mask_rcnn} further extends a mask predictor by adding a extra branch in parallel with the bounding box recognition. RetinaNet~\cite{focal_loss} is another FPN based single stage detector, which involves Focal-Loss to address class imbalance issue caused by extreme foreground-background ratio.

\textbf{Speed perspective:} 
Object detection literatures have been also strive to improve the speed of the detectors. Back to original R-CNN, which forwards each proposal separately through the whole network. He \emph{et al.} proposes SPP-net \cite{sppnet} to share the computations among candidate boxes. Both Fast/Faster R-CNN~\cite{fast_rcnn,faster_rcnn} accelerates the network by uniforming the detection pipeline. R-FCN~\cite{rfcn} shares computations between RoI subnetworks, which speed up inference when a large number of proposals are utilized. Another hot research topic is proposal free detector. YOLO and YOLO v2~\cite{yolo,yolo9000} simplifies object detection as a regression problem, which directly predicts the bounding boxes and associated class probabilities without proposal generation. SSD~\cite{ssd} further improves performance by producing predictions of different scales from different layers. Unlike box-center based detectors, DeNet~\cite{denet} first predicts all boxes corners, and then quickly searching the corner distribution for non-trivial bounding boxes.

In conclusion, from accuracy perspective, both one and two-stage detectors have achieved state-of-art precision with nearly speed. However from speed perspective, object detection literature is lack of competitive fast two-stage detector compared single-stage approaches with nearly accuracy. In this paper, we try to design a better and faster two-stage detector called Light head R-CNN to fill this missing.

\section{Our Approach} \label{sec:our_approach}
\begin{figure*}[t!]
   \begin{center}
      \includegraphics[clip=true, height=0.7\linewidth, width=0.8\linewidth]{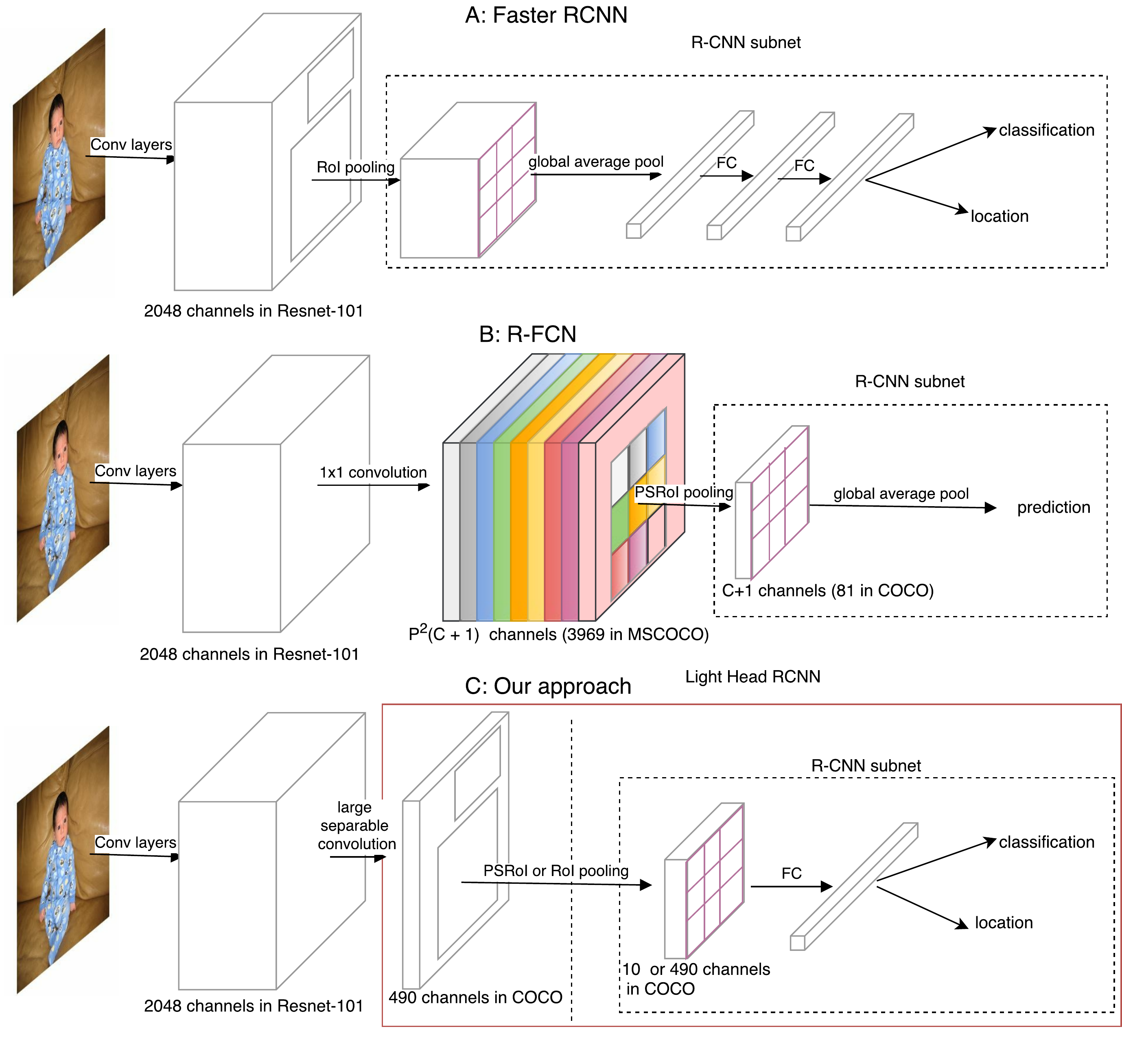}
   \end{center}
   \caption{Overview of our approach. Our Light-Head R-CNN builds ``thin" feature maps before RoI warping, by large separable convolution. We adopt a single fully-connected layer with 2048 channels in our R-CNN subnet. Thanks for thinner feature maps and cheap R-CNN subnet, the whole network is highly efficient while keeping accuracy.} 
   \label{fig:wholepipe}
\end{figure*}

In this section, we will first present our light-head R-CNN and then describe other design details in object detection.

\subsection{Light-Head R-CNN}\label{sec:light-head rcnn}
As we discuss in Section~\ref{sec:introduction}, conventional two-stage object detectors usually involve a heavy head, which has negative influence on the computational speed. ``Head'' in our paper refers to the structure attached to our backbone base network. More specifically, there will be two components: R-CNN subnet and ROI warping.

\subsubsection{R-CNN subnet}\label{sec:rcnnsubnet}
Faster R-CNN adopts a powerful R-CNN which utilizes two large fully connected layers or whole Resnet stage 5~\cite{faster_rcnn, NOC} as a second stage classifier, which is beneficial to the detection performance. Therefore Faster R-CNN and its extensions perform leading accuracy in the most challenging benchmarks like COCO. However, the computation could be intensive especially when the number of object proposals is large. To speed up RoI-wise subnet, R-FCN first produces a set of score maps for each region, whose channel number will be $\# classes \times p \times p$~($p$ is the followed pooling size), and then pool along each RoI and average vote the final prediction. Using a computation-free R-CNN subnet, R-FCN gets comparable results by involving more computation on RoI shared score maps generation.

As mentioned above, Faster R-CNN and R-FCN have heavy head but at different positions. From \textbf{accuracy} perspective, although Faster R-CNN is good at RoI classification, it usually involves global average pooling in order to reduce the computation of first fully connected layer, which is harmful for spatial localization. For R-FCN, it directly pools the prediction results after the position-sensitive pooling and the performance is usually not as strong as Faster R-CNN without RoI-wise computation layers. From \textbf{speed} perspective, Faster R-CNN pass every RoI independently through a costly R-CNN subnet, which slows down the network speed especially when the number of proposals is large. R-FCN uses cost-free R-CNN subnet as a second stage detector. But as R-FCN needs to produce a very large score map for RoI pooling, the whole network still time/memory consuming. 

Having these issues in mind, in our new Light-Head R-CNN, we propose to utilize a simple, cheap fully-connected layer for our R-CNN subnet, which makes a good trade-off between the performance and computational speed. Figure~\ref{fig:wholepipe} (C) provides the overview of our Light Head R-CNN. As the computation and memory cost for the fully-connected layer also depends on the number channel maps after ROI operation, we next discuss how we design the ROI warping.

\subsubsection{Thin feature maps for RoI warping}\label{sec:scoremap}  

Before feeding proposals into R-CNN subnet, RoI warping is involved to make the shape of feature maps fixed. 

In our Light-Head R-CNN, we propose to generate the feature maps with small channel number~(thin feature maps), followed by conventional RoI warping.  In our experiments, we find that RoI warping on \textbf{thin feature maps} will not only improves the accuracy but also saves memory and computation during training and inference. Considering PSRoI pooling on thin feature maps, we can bring more computation to strengthen R-CNN and decrease the channels. In addition, if we apply RoI pooling on our thin feature maps, we can reduce R-CNN overhead and abandon Global Average Pooling to improve performance simultaneously. Moreover, without losing time efficiency, large convolution can be involved for thin feature map production.

\subsection{Light-Head R-CNN for Object Detection}\label{sec:lightheadod} 
Following the above discussions, we present our implementation details for general object detection. The pipeline of our approach is in Figure~\ref{fig:wholepipe} (C). We have two settings: 1) setting ``L" to validate the performance our algorithm when integrated with a large backbone network; 2) setting ``S" to validate the effectiveness and efficiency of our algorithm when uses a small backbone network. Unless otherwise specified, Setting L and Setting S share the same other settings.

\noindent {\bf Basic feature extractor.}
For the setting L, we adopt ResNet 101~\cite{he2016deep} as our basic feature extractor. On the other hand, we utilize the Xception-like small base model for the setting S. The network structure of the Xception model can be found in Table~\ref{table:xception_like}. In Figure~\ref{fig:wholepipe}, ``Conv layers'' indicate our base-model. The last convolution blocks of conv4 and conv5 are denoted as $C_4, C_5$.

\noindent {\bf Thin feature maps.}
We apply large separable convolution layers~{\cite{googlenetv2,peng2017large} on $C_5$, the structure is shown in Figure~\ref{fig:large_kernel} C. In our approach, we let $k$ to 15, $C_{mid}=64$ for setting S, and $C_{mid}=256$ for setting L.  We also reduce the $C_{out}$ to $10 \times p \times p$ which is extremely small compared with limited $\# classes \times p \times p$ used in R-FCN. Benefit from the larger valid receptive field caused by large kernel, the feature maps we pooled on are more powerful.

\begin{figure}[th]
   \begin{center}
      \includegraphics[clip=true, height=0.5\linewidth, width=0.6\linewidth]{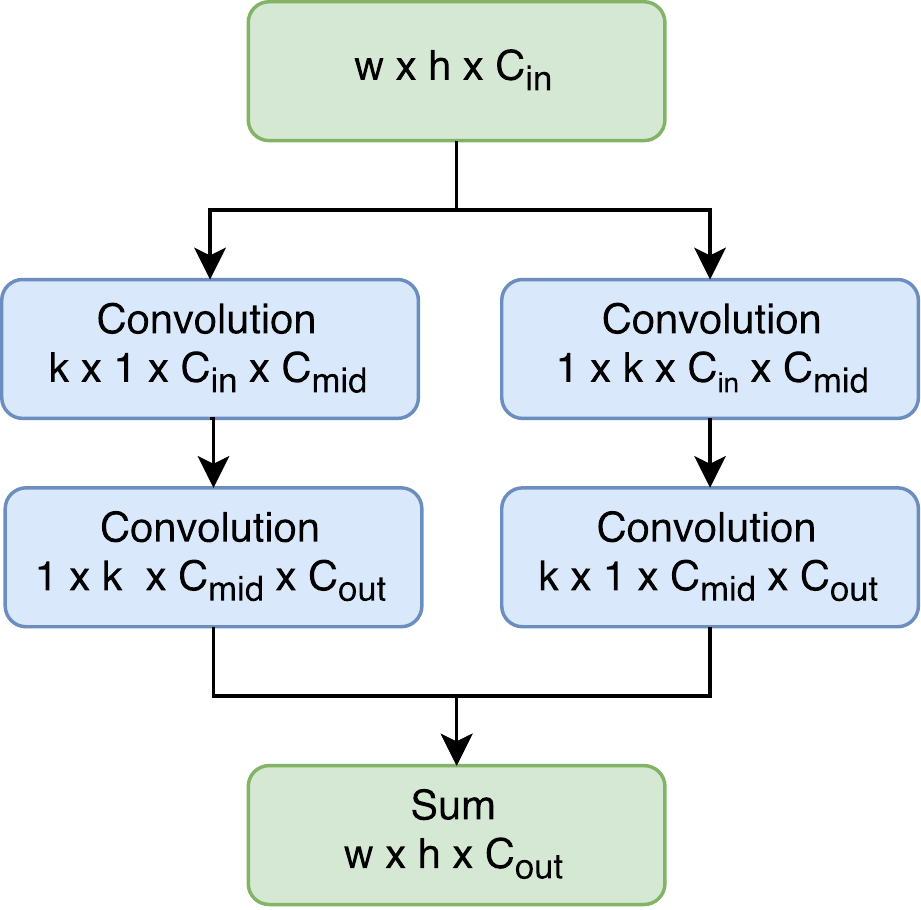}
   \end{center}
   \caption{Large separable convolution performs a $k\times 1$ and $1\times k$ convolution sequentially. The computational complexity can be further controlled through $C_{mid}, C_{out}$.}
   \label{fig:large_kernel}
\end{figure}

\noindent {\bf R-CNN subnet.} 
Here we only employ a single fully-connected layer with 2048 channels~(no dropout) in R-CNN subnet, followed by two sibling fully connected layer to predict RoI classification and regression. Only 4 channels are applied for each bounding box location because we share the regression between different classes. Benefited from the powerful feature maps for RoI warping, a simple Light-Head R-CNN can also achieve remarkable results, while keeping the efficiency.  

\noindent {\bf RPN}~(Region Proposal Network) is a sliding-window class-agnostic object detector that use features from  $C_4$. RPN pre-defines a set of anchors, which are controlled by several specific scales and aspect ratios. In our models, we set three aspect ratios \{1:2, 1:1, 2:1\} and five scales \{$32^2, 64^2, 128^2, 256^2, 512^2$\} to cover objects of different shapes. Since there are many proposals heavily overlapping with each other, non-maximum suppression~(NMS) is used to reduce the number of proposals. Before feeding them into RoI prediction subnetwork. We set the intersection-over-union~(IoU) threshold of 0.7 for NMS. Then we assign anchors training labels based on their IoU ratios with ground-truth bounding boxes. If the anchor has IoU over 0.7 with any ground-truth box, it will be set a positive label. Anchors which have highest IoU for ground-truth box will also be assigned a positive label. Meanwhile, if extra anchors have IoU less than 0.3 with all ground-truth box, their labels will be negative. More details can be referred to~\cite{faster_rcnn}.

\begin{figure*}[th]
   \begin{center}
      \includegraphics[clip=true, height=0.2\linewidth, width=0.8\linewidth]{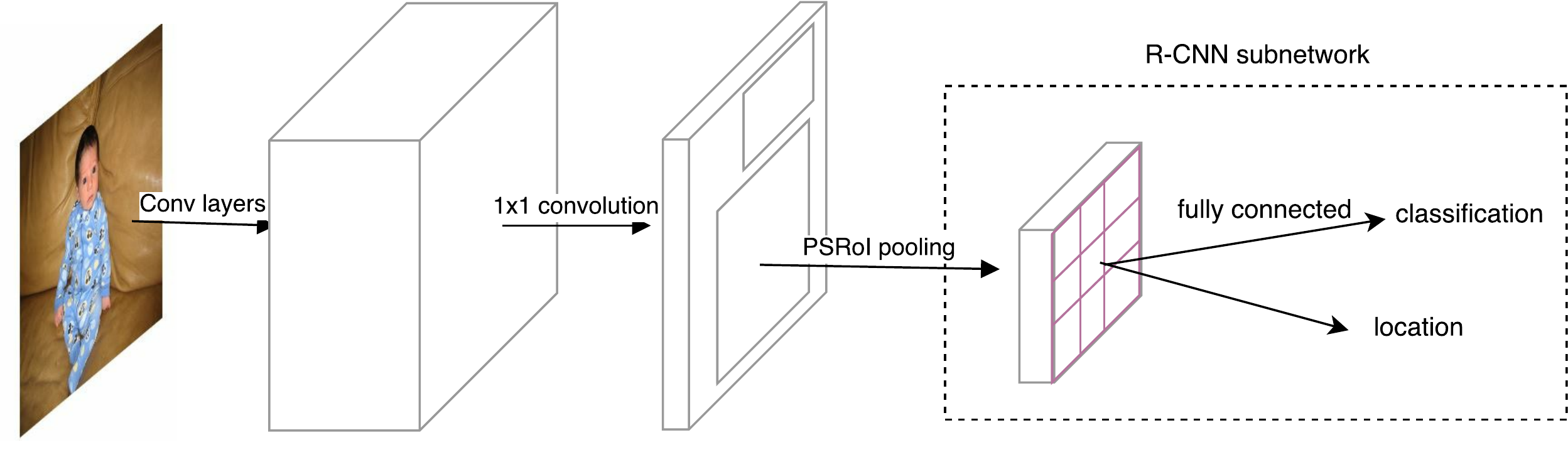}
   \end{center}
   \caption{The network to evaluate the impact of thin feature maps.  We keep the networks same as R-FCN except that we decrease the feature map channels used for PSRoI pooling. And we add additional fully-connected layers for final prediction.}
   \label{fig:less_channel}
\end{figure*}

\section{Experiments}\label{sec:experiment}   
In this section, we evaluate our approach on COCO~\cite{msCOCO,coco_api} dataset, which has 80 object categories. There are 80k \emph{train} set and 40k \emph{validation} set, which will be further split into 35k \emph{val-minusmini} and 5k \emph{mini-validation} set. Following the common setting, we combine the \emph{train} set and the \emph{val-minusmini} to obtain the 115K images for training use the 5K \emph{mini-validation} images for validation.

\subsection{Implementation Details} 
Our detector is end-to-end trained based on 8 Pascal TITAN XP GPUs using synchronized SGD with a weight decay of 0.0001 and momentum of 0.9.  Each mini-batch has 2 images per GPU and each image has 2000/1000 RoIs for training/testing. We pad images within mini-batch to the same size by filling zeros into the right-bottom of the image. Learning rate is set to 0.01 for first 1.5M iterations~(passing 1 image will be regarded as 1 iteration), and 0.001 for later 0.5M iterations.  

All experiments adopt $\grave{a}$trous~\cite{wavelet1999,fcn,fcn_crf} algorithm in stage 5 of Resnet and online hard example mining~(OHEM)~\cite{OHEM} techniques. \c{Unless explicitly stated,} our backbone network is initialized based on the pre-trained ImageNet~\cite{imagenet2015} base model and pooling size is set to 7. We fix the parameters of stage 1 and 2 in base model, batch normalization is also fixed for faster experiment. Horizontal image flipping augmentation is adopted unless otherwise noted.

In the following discussion, we will first perform a series of ablation experiments to validate the effectiveness of our method. Later, we will present the comparison with the state-of-art detectors, such as FPN~\cite{fpn}, Mask R-CNN\cite{mask_rcnn}, RetinaNet\cite{focal_loss}, on the COCO \emph{test} dataset. 


\subsection{Ablation Experiments} 

In order to fairly compare with the existing methods, we adopt Resnet101 as our backbone network for our ablation studies, similar to the Setting L described in Section\ref{sec:lightheadod}.


\subsubsection{Baselines}\label{sec:baselineexp}

Following the detail setting provided by the publicly available R-FCN code\footnote{\url{https://github.com/msracver/Deformable-ConvNets}}, we first evaluate R-FCN in our experiment, denoted as \emph{B1}, and it achieves 32.1\% mmAP in COCO \emph{mini-validation} set.

By revising the base setting, we can obtain a stronger baseline, denoted as \emph{B2}, with the following differences:
\begin{enumerate*}[label=(\roman*)]
   \item We resize the shorter edge of image to 800 pixels, and restrict the max size of the longer edge to 1200. We set 5 anchors \{$32^2, 64^2, 128^2, 256^2, 512^2$\} for RPN because of the larger image. 
   \item We find that regression loss is definitely smaller that classification loss in R-CNN. Thus we double the R-CNN regression loss to balance multi-task training. 
   \item We select 256 samples ranked based on the loss for backpropagation. We use 2000 RoIs per image for training and 1000 RoIs for testing. 
\end{enumerate*} As table~\ref{table:Baselines} shows, our method improves the mmAP by nearly 3 points.

\begin{table}[ht]
\begin{center}
\begin{tabular}{|l|c|c|c|c|}
\hline
Models & mAP & $\text{AP}_s$ & $\text{AP}_m$ & $\text{AP}_l$   \\ 
\hline 
\emph{B1} & 32.1 & 12.8 & 34.9 & 46.1 \\
\emph{B2} & 35.2  & 19.1 & 39.6 & 46.5 \\
\hline
\end{tabular}
\end{center}
\caption{Baselines of our proposed methods. Baseline \emph{B1} is the original R-FCN. Baseline \emph{B2} involves our reproduced implementation details, such as more balanced loss and large image size.}
\label{table:Baselines}
\end{table}

\subsubsection{Thin feature maps for RoI warping}
We investigate the impact of reducing channels of feature maps for ROI warping following the baseline setting described in Section~\ref{sec:baselineexp}.  To achieve this goal, we design a simple network structure which is shown in Figure~\ref{fig:less_channel}. The whole pipeline is exactly same as original R-FCN for fairly comparison. Except the following small difference: 
\begin{enumerate*}[label=(\roman*)]
   \item We reduce the feature map channels to 490~($10 \times 7 \times 7$) for PSRoI pooling. Noticing it is quite different from the original R-FCN, which involves 3969~($81 \times 7 \times 7$) channels.
   \item As we modify the channel number of feature maps, we can't directly vote for the final predictions. A simple fully connected layer is involved for final prediction.
\end{enumerate*} 

Results are shown in table~\ref{table:reduce_channels}. Although the channel number has been largely decrease, 3969 vs 490, our have comparable performance. In addition, it is important to note that, with our Light-head R-CNN design, it enables us to efficiently integrate feature pyramid network~\cite{fpn} as shown in Table~\ref{table:COCO_methods_comparison}. This is almost impossible for the original R-FCN as the memory consumption will be high if we want to perform position-sensitive pooling on the high-resolution feature maps like Conv2~(Resnet stage 2). 

To validate the influence of reducing channels in Faster R-CNN, we also try to replace the PSRoI pooling with conventional RoI pooling, it slightly improves the accuracy with 0.3 gain. One hypothesis is that RoI pooling involves more features~(49x) in second stage, the accuracy gain is benefited from more computation.

\begin{table}[ht]
\begin{center}
\begin{tabular}{|l|c|c|c|c|}
\hline
Models & mmAP & $\text{AP}_s$ & $\text{AP}_m$ & $\text{AP}_l$   \\ 
\hline 
\emph{B1} & 32.1 & 12.8 & 34.9 & 46.1 \\
thin feature maps \emph{B1} & 31.4  & 12.2 & 35.5 & 47.4 \\
\emph{B2} & 35.2  & 19.1 & 39.6 & 46.5 \\
thin feature maps \emph{B2} & 34.6  & 17.6 & 39.1 & 48.0 \\
\hline
\end{tabular}
\end{center}
\caption{The impact of reducing feature map channels for RoI warping. We show the comparison between original R-FCN and R-FCN with thin feature maps for RoI warping. Also we show the influence in our stronger baseline. We use mmAP to indicate results of mAP@[0.5:0.95].}
\label{table:reduce_channels}
\end{table}

\paragraph{Large separable convolution} 

The feature map channels for RoI warping is pretty small in our design. In original implementation, $1 \times 1$ convolution is involved to reshape a small channel, which decrease feature map capability. We involve large separable convolution to enhance these feature maps while keeping small channels. We show the structure of large kernel in Figure~\ref{fig:large_kernel}. The hyperparameter we set is $k = 15, C_{mid} = 256, C_{out} = 490$. In table~\ref{table:powerful_fm},  compared with the results based on our reproduced R-FCN setting \emph{B2}, the thin feature map produced by large kernel can improve the performance by 0.7 points. 

\begin{table}[ht]
\begin{center}
\begin{tabular}{|l|c|c|c|c|}
\hline
Models & mmAP & $\text{AP}_s$ & $\text{AP}_m$ & $\text{AP}_l$   \\ 
\hline 
\emph{B2} & 35.2  & 19.1 & 39.6 & 46.5 \\
\hline 
+Large Kernel & 35.9 &19.2 & 40.4 & 48.3\\
\hline 
\end{tabular}
\end{center}
\caption{The impact of enhance thin feature map for RoI warping. \emph{B2} is our implemented strong R-FCN baseline.}
\label{table:powerful_fm}
\end{table}


\subsubsection{R-CNN subnet}
Here we evaluate the impact of Light version of R-CNN in R-CNN subnet as shown in Figure~\ref{sec:our_approach}. A single fully connected layer~(without dropout) with 2048 channels is employed in RoI subnetwork. Since the feature map we pooled on is small in channels~(10 in our experiments), Light R-CNN is extremely fast for per-region classification and regression. The comparison of the light-RCNN along with our reproduced stronger Faster R-CNN and R-FCN (\emph{B2} is shown in Table~\ref{table:light_RCNN}.

Our methods achieve 37.7 mmAP when combined large kernel feature maps and Light RCNN. In our experiments, under same basic settings, Faster R-CNN and R-FCN get results of 35.5/35.1 mmAP, which is much lower than our methods. More importantly, because of extremely thin feature maps and light R-CNN, we keep the time efficiency even thousands of proposals are utilized.

\begin{table}[ht]
\begin{center}
\begin{tabular}{|l|c|}
\hline
Models & mAP@[0.5:0.95]   \\ 
\hline 
baseline Faster R-CNN & 35.6 \\
\hline 
\emph{B2} (R-FCN) & 35.2\\
\hline 
+Large Kernel & 35.9 \\
+Large Kernel and Light-Head R-CNN & 37.7  \\
\hline 
\end{tabular}
\end{center}
\caption{The effectivenss of Light-Head R-CNN. The baselines of R-FCN and Fast R-CNN based on our setting L~(\ref{sec:lightheadod}).}
\label{table:light_RCNN}
\end{table}

\begin{table*}[h]
\begin{center}\small
\begin{tabular}{l|c|c|c|c|c|c|c}
\hline
method & \begin{tabular}{@{}c@{}}test size \\ shorter edge/max size\end{tabular} &  $\begin{tabular}{@{}c@{}}feature \\ pyramid \end{tabular}  $ & align  & mAP@[0.5:0.95] & $\text{AP}_s$ & $\text{AP}_m$ & $\text{AP}_l$  \\
\hline
R-FCN~\cite{rfcn} & 600/1000  & &  & 32.1 & 12.8 &34.9&46.1 \\
Faster R-CNN~(2fc) & 600/1000  &  & &  30.3  & 9.9 & 32.2 & 47.4  \\
Deformable~\cite{deformable} & 600/1000   & &  $\surd$ & 34.5 & 14.0 & 37.7 & 50.3 \\
G-RMI~\cite{GRMI} & 600/1000 & & &  35.6 & - & - & - \\
FPN~\cite{fpn} & 800/1200 &  $\surd$&  & 36.2 & 18.2 & 39.0 & 48.2\\
Mask R-CNN~\cite{mask_rcnn} & 800/1200     & $\surd$ & $\surd$& 38.2 & 20.1 & 41.1 &50.2 \\
RetinaNet~\cite{focal_loss} & 800/1200  & $\surd$ & & 37.8 & 20.2 &  41.1 & 49.2 \\
RetinaNet ms-train~\cite{focal_loss} & 800/1200  & $\surd$ & & 39.1 & 21.8 &  42.7 & 50.2 \\
\hline
Light head R-CNN & 800/1200   & & $\surd$ & \textbf{39.5}  & 21.8 & 43.0 &  50.7 \\
Light head R-CNN ms-train& 800/1200   & & $\surd$ & \textbf{40.8} & 22.7 & 44.3 &  52.8 \\
Light head R-CNN & 800/1200   & $\surd$ & $\surd$ & \textbf{41.5} & 25.2 & 45.3 &  53.1 \\
\hline
\end{tabular}
\end{center}
\caption{Comparisons of single size tested single-model results on COCO test-dev. All experiments are using Resnet-101 as basic feature extractor~(except R-RMI with Inception Resnet V2~\cite{inception-resv2}). Light-Head R-CNN reaches a new state-of-the-art accuracy. Noticing results of test-dev is slightly different from mini-validation. ``ms-train" means multi-scale training.}
\label{table:COCO_methods_comparison}
\end{table*}

\begin{table}[ht]
\begin{center}
\begin{tabular}{|l|c|}
\hline
Models & mAP@[0.5:0.95]   \\ 
\hline 
Light-Head R-CNN & 37.7  \\
\hline 
+ pool with alignment & 39.0 \\
+ NMS with 0.5 threshold & 39.6 \\
+ multiscale train & 40.6 \\
\hline 
\end{tabular}
\end{center}
\caption{Further improvements for our approach. Results are evaluated on COCO mini-validation set.}
\label{table:further_improve}
\end{table}


\begin{table}[th]
\centering
\begin{tabular}{lcccccccc}
\hline
Layer & Output size & KSize & S & Rep & \begin{tabular}{@{}c@{}}output \\ channels \end{tabular} \\
\hline
Image & $224\times 224$ & & &  \\
\hline

Conv1 & $112\times 112$ & $3\times 3$ & 2 & 1 & 24 & \\
MaxPool & $56\times 56$ & $3\times 3$ & 2 &  &  \\
\hline
Stage2 & $28\times 28$ & & 2 & 1 & 144  \\
 & $28\times 28$ & & 1 & 3 & 144 \\
\hline
Stage3 & $14\times 14$ & & 2 & 1 & 288 &  \\
 & $14\times 14$ & & 1 & 7 & 288 &  \\
\hline
Stage4 & $7\times 7$ & & 2 & 1 & 576  \\
& $7\times 7$ & & 1 & 3 & 576 \\
\hline
GAP & $1\times1$ & $7\times 7$ & & &  \\ 
\hline
FC & & & & & 1000  \\
\hline
\hline
Comp* & & & & & 145M \\
\hline
\end{tabular}
\caption{An efficient Xception like architecture for our fast detector. Comp* indicates the complexity of the networks~(FLOPs)}
\label{table:xception_like}
\end{table}

\begin{table}[h]
\centering
\small
\begin{tabular}{l|c|c|c|c}
\hline
Model & backbone & \begin{tabular}{@{}c@{}}test size \\ short/max \\ edge \end{tabular} & \begin{tabular}{@{}c@{}}speed \\ (fps) \end{tabular} & \begin{tabular}{@{}c@{}}mAP \\ @[0.5:0.95] \end{tabular} \\ 
\hline
YOLO V2 & Darknet & 448/448 & 40 & 21.6 \\
SSD & VGG & 300/300 & 58 & 25.1 \\ 
SSD & Resnet101 & 300/300 & 16 & 28.0 \\ 
SSD & Resnet101 & 500/500 & 8 & 31.2 \\
DSSD~\cite{dssd} & Resnet101 &  300/300 & 8 & 28.0 \\ 
DSSD & Resnet101 &  500/500 & 6 & 33.2 \\ 
R-FCN & Resnet101 & 600/1000 & 11 & 29.9 \\
DeNet & Resnet34 &  512/512& 83 & 29.4 \\
\hline
 \begin{tabular}{@{}c@{}}Light Head \\ R-CNN \end{tabular} &xception* & 800/1200 & \textbf{95} & \textbf{31.5} \\
\begin{tabular}{@{}c@{}}Light Head \\ R-CNN \end{tabular}  &xception* & 700/1100 & \textbf{102} & \textbf{30.7} \\
\hline
\end{tabular}
\caption{Comparisons of our fast detector results on COCO test-dev. Xception* is a small xception like backbone. By involving a tiny base-model, Light R-CNN achieves superior performance on both accuracy and speed, which shows the flexiblilty of our proposed design.}
\label{table:speed_comparison_full}
\end{table}

\subsection{Light-Head R-CNN: High Accuracy}  

To combine with state-of-art object detectors, we following the Setting L described in Section~\ref{sec:lightheadod}. In addition, we involve interpolation techniques for PSRoI pooling which is proposed in RoIAlign~\cite{mask_rcnn}. Results are shown in table~\ref{table:further_improve}. It can bring 1.3 points gain. We also apply the common used scale jitter approach. During our training, we randomly sample the scale from \{600, 700, 800, 900, 1000\} pixels, then resize the shorter edge of the image into the sampled scale. The max edge of the image is constrained within 1400 pixels because the shorter edge may reach to 1000 pixels. Training time is also increased due to the data augmentation. Multi-scale training brings nearly 1 points improvement on mmAP. We finally replace original 0.3 threshold with 0.5 for Non-maximum Suppression (NMS). It improves 0.6 points of mmAP by improving the recall rate especially for the crowd cases.

Table~\ref{table:COCO_methods_comparison} also summarizes the results from state-of-the-art detectors including both one-stage and two-stage methods on COCO test-dev datasets. Our single scale tested model can achieve 40.8\% mmAP without bells and whistles, significantly surpassing all the competitors~(40.8 vs 39.1). It validates that our Light Head R-CNN would be a good choice for the large backbone models and promising results can be obtained without heavy computation cost. Some of the illustrative results are shown in Figure~\ref{fir:accu_model}.

\subsection{Light-Head R-CNN: High Speed}
Larger backbone like Resnet 101 is slow based on the computational speed. To validate the efficiency of Light-Head R-CNN, we produce an efficient bottleneck xception like network with 34.1 top 1 error~(at $224 \times 224$  in ImageNet) to evaluate our methods. The backbone network architecture is shown in Table~\ref{table:xception_like}. Following xception design strategies, we replace all convolution layer in bottle-neck structure with channel-wise convolution. However we do not use the pre-activation design which is proposed in identity mappings~\cite{identity_mappings} because of shallow network.  The implementation details for our Light Head R-CNN can be found from the Setting S in Section~\ref{sec:lightheadod}.

More specifically, we have the following changes for the fast inference speed:
\begin{enumerate*}[label=(\roman*)]
\item We replace Resnet-101 backbone in Setting L with a tiny xception like network.
\item We abandon $\grave{a}$trous algorithm in our fast models, because it involves much computation compared with small backbone.
\item we set RPN convolution to 256 channels, which is half of original used in Faster R-CNN and R-FCN.
\item we apply large separable convolution with $kernel\_size = 15, C_{mid} = 64, C_{out} = 490$~($10 \times 7 \times 7$). Since the middle channels we used is extremely small, large kernel is still efficient for inference.
\item We adopt PSPooling with alignment technique as our RoI warping, since it reduces the pooled feature map channels by $k \times k$ times~($k$ is the pooling size). Noticing that it will obtain better results if we involve RoI-align.
\end{enumerate*} 

We compared our methods with recent fast detectors like YOLO, SSD, and DeNet. Results are evaluated on COCO test-dev datasets, only one batch is adopted and we absorb the Batch-normalization~\cite{bath_norm} for fast inference. As shown in Table~\ref{table:speed_comparison_full}, our methods get 30.7 mmAP at 102 FPS on MS COCO, significantly outperforming the fast detectors like YOLO and SSD. Some of the results are visualized in Figure~\ref{fir:fast_model}.

\section{Conclusion}
In this paper, we present Light Head R-CNN, which involves a better design principle for two-stage object detectors. Compared with traditional two-stage detectors like Faster R-CNN and R-FCN, which usually have a heavy head, our light head design enables us to significantly improve the detection results, without compromising the computational speed. More importantly, compared with the fast single-stage detectors like YOLO and SSD, we obtain superior performance even with faster computational speed. For example, our Light Head R-CNN coupled with small Xception-like base model can achieve 30.7 mmAP at the speed of 102 FPS. 

\begin{figure*}[ht]
   \begin{center}
      \includegraphics[clip=true, height=0.43\linewidth,width=1\linewidth]{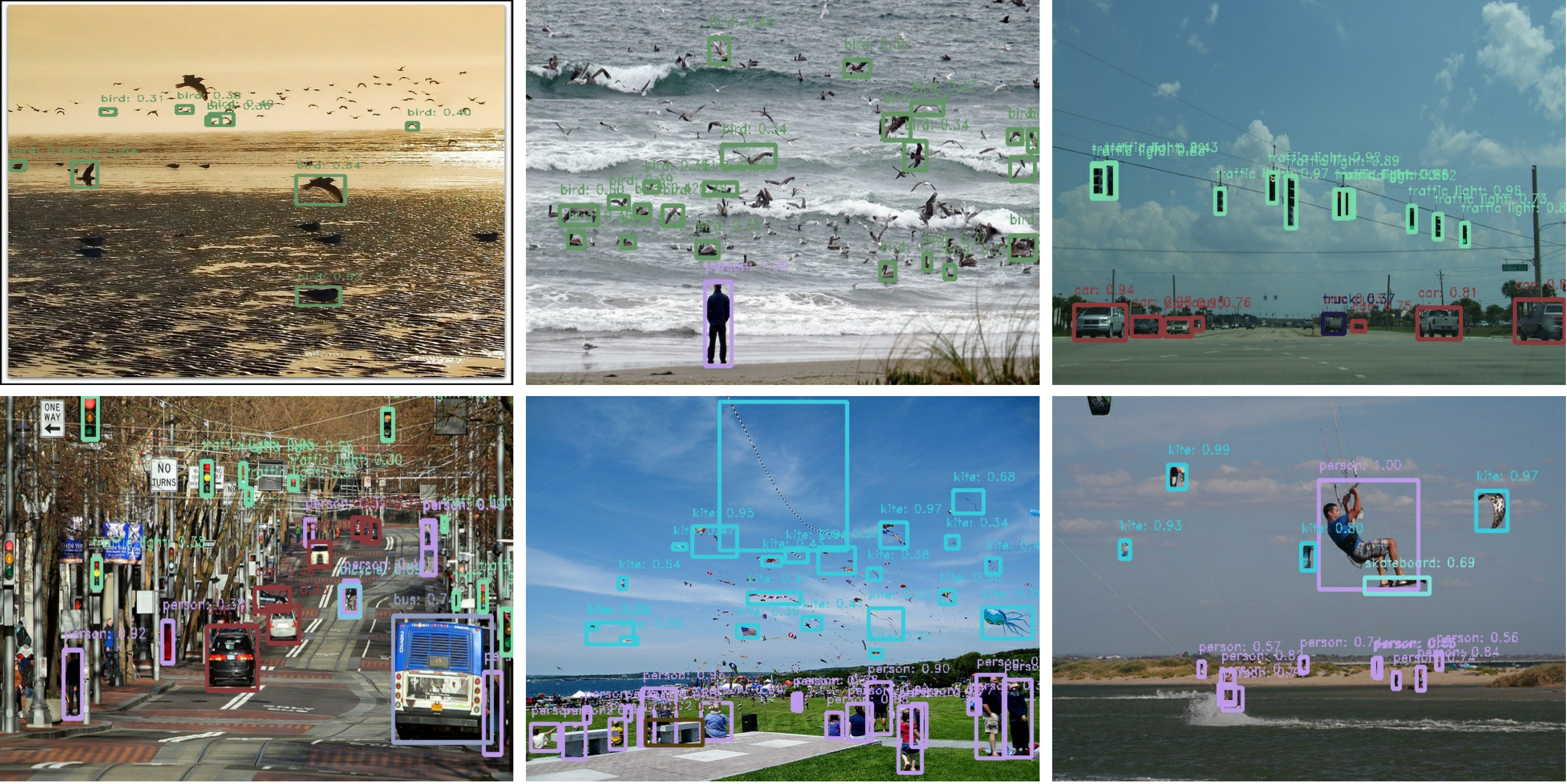}
   \end{center}
   \caption{Representative results of our large ``L" model.  }
   \label{fir:accu_model}
\end{figure*}

\begin{figure*}[h]
   \begin{center}
      \includegraphics[clip=true, height=0.43\linewidth ,width=1\linewidth]{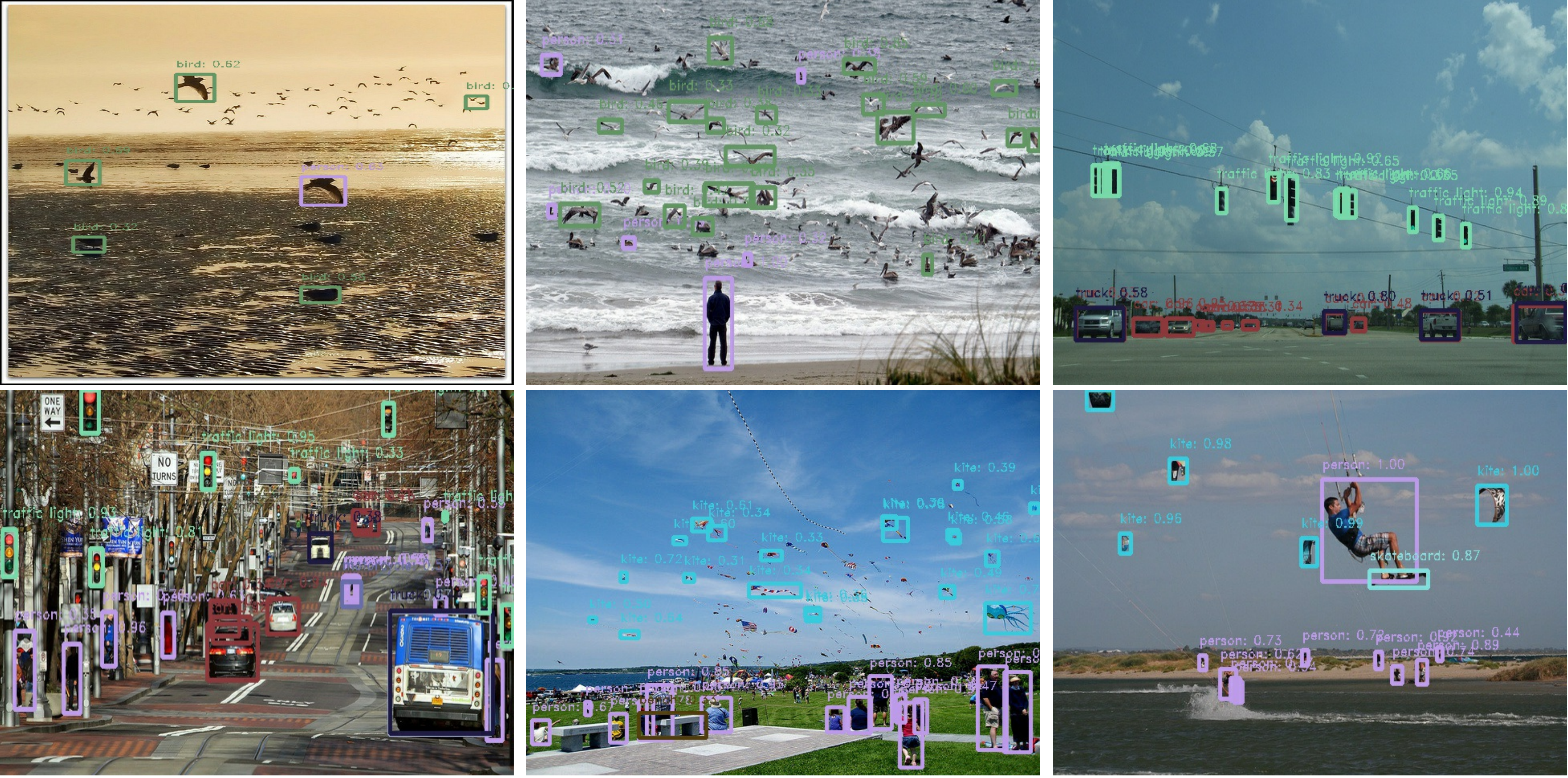}
   \end{center}
   \caption{Representative results of our small ``S" model.}
   \label{fir:fast_model}
\end{figure*}

{\small
\bibliographystyle{ieee}
\bibliography{egbib}
}

\end{document}